%% file: main.tex
\documentclass[a4paper,twoside]{article}
\usepackage{epsfig}
\usepackage{subcaption}
\usepackage{calc}
\usepackage{amssymb}
\usepackage{amstext}
\usepackage{amsmath}
\usepackage{mathtools}
\usepackage{amsthm}
\usepackage{multicol}
\usepackage{pslatex}
\usepackage{natbib}
\usepackage[bottom]{footmisc}
\usepackage{placeins}
\usepackage{tikz}
\usepackage{amsmath}
\usetikzlibrary{automata}
\DeclareMathOperator*{\argmax}{argmax} 
\newcommand{\FGSM}{\text{FGSM}}

\usepackage{booktabs}

\usepackage[colorlinks=true, linkcolor=blue, citecolor=blue]{hyperref}
\usepackage[capitalise,noabbrev,nameinlink]{cleveref}

\usepackage{SCITEPRESS}     
\usepackage{algorithm}
\usepackage{algpseudocode}

\usepackage{amsthm}

\theoremstyle{definition}
\newtheorem{definition}{Definition}[section]

\usetikzlibrary{shapes,shadows,arrows,positioning,angles,quotes}
\tikzset{My Arrow Style/.style={single arrow, fill=black!15, anchor=base, align=center,text width=2.3cm}}
\tikzstyle{arrow} = [thick,->,>=stealth]
\usetikzlibrary{calc,trees,positioning,arrows,fit,shapes,calc}

\tikzstyle{startstop} = [rectangle, rounded corners, minimum width=1.5cm, minimum height=0.5cm,text centered, draw=black, fill=red!30]
\tikzstyle{io} = [trapezium, trapezium left angle=70, trapezium right angle=110, minimum width=1cm, minimum height=0.5cm, text centered, draw=black, fill=blue!30]

\tikzstyle{process} = [rectangle, minimum width=3cm, minimum height=0.5cm, text centered, draw=black, fill=orange!30]
\tikzstyle{decision} = [diamond, minimum width=1cm, minimum height=0.5cm, text centered, draw=black, fill=green!30]

\tikzstyle{arrow} = [thick,->,>=stealth]

\DeclarePairedDelimiter{\norm}{\lVert}{\rVert}
\newcommand{\scaleTable}[1]{\scalebox{0.875}{#1}}

\begin{document}

\title{Targeted Adversarial Attacks on Deep Reinforcement Learning Policies via Model~Checking}


\author{\authorname{Dennis Gross\sup{1}, Thiago D. Sim{\~a}o\sup{1}, Nils Jansen\sup{1}, and Guillermo A. P{\'e}rez\sup{2}}
\affiliation{\sup{1}Institute for Computing and Information Sciences, Radboud University, Toernooiveld 212, 6525 EC Nijmegen, \\The Netherlands}
\affiliation{\sup{2}Department of Computing, University of Antwerp, Middelheimlaan 1, 2020 Antwerpen, Belgium}
\email{dgross@science.ru.nl}
}


\abstract{
Deep Reinforcement Learning (RL) agents are susceptible to adversarial noise in their observations that can mislead their policies and decrease their performance.
However, an adversary may be interested not only in decreasing the reward, but also in modifying specific temporal logic properties of the policy.
This paper presents a metric that measures the exact impact of adversarial attacks against such properties.
We use this metric to craft optimal adversarial attacks.
Furthermore, we introduce a model checking method that allows us to verify the robustness of RL policies against adversarial attacks.
Our empirical analysis confirms (1) the quality of our metric to craft adversarial attacks against temporal logic properties, and (2) that we are able to concisely assess a system's robustness against attacks.}

\onecolumn \maketitle \normalsize \setcounter{footnote}{0} \vfill

\section{\uppercase{Introduction}}
\emph{Deep reinforcement learning (RL)} has changed how we build agents for sequential decision-making problems \citep{mnih2015human,levine2016end}.
It has triggered applications in critical domains like energy, transportation, and defense~\citep{farazi2021deep,nakabi2021deep,boron2020developing}.
An RL agent learns a near-optimal policy (based on a given objective) by making observations and gaining rewards through interacting with the environment~\citep{sutton2018reinforcement}.
Despite the success of RL, potential security risks limit its usage in real-life applications.
The so-called adversarial attacks introduce noise into the observations and mislead the RL decision-making to drop the cumulative reward, which may lead to unsafe behaviour~\citep{DBLP:conf/iclr/HuangPGDA17,DBLP:journals/cybersec/ChenLXNTH19,DBLP:journals/tai/IlahiUQJAHN22,moos2022robust,amodei2016concrete}.

Generally, rewards lack the expressiveness to encode complex safety requirements \citep{DBLP:journals/aamas/VamplewSKRRRHHM22,DBLP:conf/formats/HasanbeigKA20}.
Therefore, for an adversary, capturing how much the cumulative reward is reduced may be too generic for attacks targeting specific safety requirements.
For instance, an RL taxi agent may be optimized to transport passengers to their destinations.
With the already existing adversarial attacks, the attacker can prevent the agent from transporting the passenger.
However, the attacker cannot create controlled adversarial attacks that may increase the probability that the passenger never gets picked up or that the passenger gets picked up but never arrives at its destination.
More generally, current adversary attacks are not able to control temporal logic properties.

This paper aims to combine adversarial RL with rigorous model checking~\citep{DBLP:books/daglib/0020348}, which allows the adversary to create so-called \emph{property impact attacks (PIAs)} that can influence specific RL policy properties.
These PIAs are not limited by properties that can be expressed by rewards~\citep{littman2017environment,DBLP:conf/tacas/HahnPSSTW19,DBLP:conf/formats/HasanbeigKA20,DBLP:journals/aamas/VamplewSKRRRHHM22}, but support a broader range of properties that can be expressed by \emph{probabilistic computation tree logic} \citep[PCTL;][]{DBLP:journals/fac/HanssonJ94}.
Our experiments show that for PCTL properties, it is possible to create targeted adversarial attacks that influence them specifically.
Furthermore, the combination of model checking and adversarial RL allows us to verify via \emph{permissive policies}~\citep{DBLP:journals/corr/DragerFK0U15} how vulnerable trained policies are against~PIAs. 
Our \emph{main contributions are:}

\begin{itemize}
    \item a metric to measure the impact of adversarial attacks on a broad range of RL policy properties,
    \item a property impact attack (PIA) to target specific properties of a trained RL policy, and
    \item a method that checks the robustness of RL policies against adversarial attacks.
\end{itemize}

\noindent
The empirical analysis shows that the method to attack RL policies can effectively modify PCTL properties.
Furthermore, the results support the theoretical claim that it is possible to model check the robustness of RL policies against property impact~attacks.

The paper is structured in the following way.
First, we summarize the related work and position our paper in it.
Second, we explain the fundamentals of our technique.
Then, we present the adversarial attack setting, define our property impact attack, and show a way to model check policy robustness against such adversarial attacks.
After that, we evaluate our methods in multiple environments.

\section{\uppercase{Related Work}}
We now summarize the related work and position our paper in between adversarial RL and model checking.

There exist a variety of adversarial attack methods to attack RL policies with the goal of dropping their total expected reward \citep{DBLP:conf/ccs/ChanWY20,DBLP:journals/corr/abs-1710-00814,DBLP:journals/tai/IlahiUQJAHN22,DBLP:conf/iclr/LinHLS0S17,DBLP:conf/trustcom/ClarkDG18,DBLP:conf/aaai/YuS22}.
The first proposed adversarial attack on deep RL policies \citep{DBLP:conf/iclr/HuangPGDA17} uses a modified version of the \emph{fast gradient sign method (FGSM)}, developed by \citet{DBLP:journals/corr/GoodfellowSS14}, to force the RL policy to make malicious decisions (for more details, see \cref{sec:aarl}).
However, none of the previous work let the attacker target temporal logic properties of RL policies.
\citet{DBLP:conf/ccs/ChanWY20} create more effective attacks that modify only one feature (if the smallest sliding window is used) of the observation of the agent, their approach empirically measures the impact of each feature in the reward, then it modifies the feature with the highest impact.
We build upon this idea and measure the impact of changing each feature in a given temporal logic property instead of the reward.

There exist a large body of work that combines RL with model checking~\citep{yuwangPCTL,DBLP:conf/formats/HasanbeigKA20,DBLP:conf/tacas/HahnPSSTW19,DBLP:conf/aiia/HasanbeigKA19,fulton2019verifiably,DBLP:conf/cdc/SadighKCSS14,bouton2019reinforcement,DBLP:conf/aaai/Chatterjee0PRZ17} but no work that uses model checking to create adversarial attacks for RL policies~\citep{DBLP:journals/cybersec/ChenLXNTH19,DBLP:journals/tai/IlahiUQJAHN22,moos2022robust}.
Most work about the formal robustness checking of deep learning models focuses on supervised learning~\citep{DBLP:conf/cav/KatzBDJK17,DBLP:conf/cav/KatzHIJLLSTWZDK19,DBLP:conf/sp/GehrMDTCV18,DBLP:conf/cav/HuangKWW17,DBLP:conf/ijcai/RuanHK18}.
In the RL setting, \citet{DBLP:conf/nips/0001CX0LBH20} introduce a formal approach to check the robustness against adversarial attacks with respect to the reward.
They formulate the perturbation on state observations as a modified \emph{Markov decision process (MDP)}.
Furthermore, they can obtain certain robustness certificates under attack.
For environments like Pong, they can guarantee actions do not change for all frames during policy execution, thus guaranteeing the cumulative rewards under attack.
We, on the other hand, focus on the robustness of temporal PCTL~properties against attacks.

\section{\uppercase{Background}}
In this section, we introduce the necessary foundations.
First, we summarize the modeling and analysis of probabilistic systems.
Second, we introduce a method to attack deep RL policies and a method that increases the robustness of trained RL policies.

\subsection{Probabilistic Systems}

A \textit{probability distribution} over a set $X$ is a function $\mu : X \rightarrow [0,1]$ with $\sum_{x \in X} \mu(x) = 1$. The set of all distributions over $X$ is denoted by $Distr(X)$.

\begin{definition}[Markov Decision Process]
A Markov decision process (MDP) is a tuple $M = (S,s_0,Act,T,rew)$ where
$S$ is a finite, nonempty set of states, $s_0 \in S$ is an initial state, $Act$ is a finite set of actions, $T\colon S \times Act \rightarrow Distr(S)$ is a probability transition function. We employ a factored state representation $S \subseteq \mathbb{Z}^n$, where each state $s\in\mathbb{Z}^n$ is an $n$-dimensional vector of features $(f_1, f_2, ...,f_n)$ such that $f_i \in \mathbb{Z}$ for $1 \leq i \leq n$. 
We define $rew \colon S \times Act \rightarrow \mathbb{R}$ as a reward~function.
\end{definition}

The available actions in $s \in S$ are $Act(s) = \{a \in Act \mid T(s,a) \neq \bot\}$.
An MDP with only one action per state ($\forall s \in S \colon |Act(s)| = 1$) is a discrete-time Markov chain (DTMC).
Note that features do not necessarily have to have the same domain size.
We define $\mathcal{F}$ as the set of all features $f_i$ in state $s \in S$.

A path of an MDP $M$ is an (in)finite sequence $\tau = s_0 \xrightarrow[\text{}]{\text{$a_0,r_0$}} s_1 \xrightarrow[\text{}]{\text{$a_1,r_1$}}...$, where $s_i \in S$, $a_i \in Act(s_i)$, $r_i \vcentcolon= rew(s_i,a_i)$, and $T(s_i,a_i)(s_{i+1}) \neq 0$.
A state $s'$ is reachable from state $s$ if there exists a path $\tau$ from state $s$ to state $s'$.
We say a state $s$ is reachable if $s$ is reachable from $s_0$.

\begin{definition}[Policy]
A memoryless deterministic policy for an MDP $M {=} (S,s_0,Act,T,rew)$ is a function $\pi \colon S \rightarrow Act$ that maps a state $s \in S$ to an action $a \in Act(s)$.
\end{definition}

Applying a policy $\pi$ to an MDP $M$ yields an \emph{induced DTMC}, denoted as $D$, where all non-determinism is resolved.
This way, we say a state $s$ is reachable by a policy $\pi$ if $s$ is reachable in the DTMC induced by $\pi$.
$\Lambda$ is the set of all possible memoryless policies.

To analyze the properties of an induced DTMC, it is necessary to specify the properties via a specification language like probabilistic computation tree logic PCTL~\citep{DBLP:journals/fac/HanssonJ94}.
\begin{definition}[PCTL Syntax]\label{def:pctl}
    Let $AP$ be a set of atomic propositions. The following grammar defines a state formula:
    $\Phi  \coloneqq \text{ true }|\text{ a }| \text{ } \Phi_1 \land \Phi_2 \text{ }|\text{ }\lnot \Phi\text{ }|P_{\bowtie p}| P^{max}_{\bowtie p}(\phi)\text{ }|\text{ }P^{min}_{\bowtie p}(\phi)$ where $a \in AP, \bowtie \in \{<,>,\leq,\geq\}$, $p \in [0,1]$ is a threshold, and $\phi$ is a path formula which is formed according to the following grammar $\phi \coloneqq X\Phi\text{ }|\text{ }\phi_1\text{ }U\text{ }\phi_2\text{ }|\text{ }\phi_1\text{ }F_{\theta t}\text{ }\phi_2\text{ }|G\text{ }\Phi$ with $\theta = \{<,\leq\}$.
\end{definition}
PCTL formulae are interpreted over the states of an induced DTMC.
In a slight abuse of notation, we use PCTL state formulas to denote probability values. That is, we sometimes write 
$P_{\bowtie p}(\phi)$ where we omit the threshold $p$. For instance,
$P(F_{\leq100} collision)$ 
denotes the reachability probability of eventually running into a collision within the first $100$ time steps.

There is a variety of model checking algorithms for verifying PCTL properties~\citep{DBLP:conf/focs/CourcoubetisY88,DBLP:journals/jacm/CourcoubetisY95},
and PRISM and Storm offer efficient and mature tool support~\citep{DBLP:conf/cav/KwiatkowskaNP11,DBLP:journals/sttt/HenselJKQV22}.
COOL-MC~\citep{DBLP:conf/setta/Gross22} allows model checking of a trained RL policy against
a PCTL property and MDP.
The tool builds the induced DTMC on the fly via an \emph{incremental building process}~\citep{DBLP:conf/concur/CassezDFLL05,DBLP:conf/tacas/DavidJLMT15}.

\subsection{Adversarial Attacks on Deep RL Policies}\label{sec:aarl}
The standard learning goal for RL is to find a policy $\pi$ in a MDP such that $\pi$ maximizes the expected accumulated discounted rewards, that is, $\mathbb{E}[\sum^{L}_{t=0}\gamma^t R_t]$, where $\gamma$ with $0 \leq \gamma \leq 1$ is the discount factor, $R_t$ is the reward at time $t$, and $L$ is the total number of steps.
Deep RL uses neural networks to train policies.
A neural network is a function parameterized by weights~$\theta$.
In deep RL, the policy $\pi$ is encoded using a neural network which can be trained by minimizing a sequence of loss functions $J(\theta, s, a)$~\citep{mnih2013playing}.

An \emph{adversary} is a malicious actor that seeks to harm or undermine the performance of an RL system.
For instance, an adversary may try to decrease the expected discounted reward by attacking the RL policy via adversarial attacks.

\begin{figure}
     \begin{subfigure}[b]{\columnwidth}
        \centering
        \begin{tikzpicture}[]
            \node (agent1) [process] {RL policy};
            \node (env) [process, below of=agent1,yshift=0.1cm,xshift=2cm] {Environment};
            \draw [arrow,] (agent1) -| node[anchor=west] {$\pi(s)$} (env);
            \draw [arrow] (env) -| node[anchor=east] {\begin{tabular}{c}$s$\\$rew$ \end{tabular}} (agent1);
        \end{tikzpicture}
        \caption{RL policy interaction with the environment.}
        \label{fig:normal_rl}
    \end{subfigure}
    \hfill \\
    \begin{subfigure}[b]{\columnwidth}
        \centering
        \begin{tikzpicture}[]
            \node (agent1) [process] {RL policy};
            \node (env) [process, below of=agent1,yshift=0.1cm,xshift=2.5cm] {Environment};
            \node (adv) [process, below of=env,yshift=0.1cm,xshift=-2.5cm] {Attacker};
            \draw [arrow, red, dashed] (agent1) -| node[anchor=west] {$\pi(s_{adv})$} (env);
            \draw [arrow] (env) |- node[anchor=west] {\begin{tabular}{c}$s$\\$rew$ \end{tabular}} (adv);
            \draw [arrow, red, dashed] (adv) -- node[anchor=east] {\begin{tabular}{c}$s_{adv}=\delta(s)$\\$rew$ \end{tabular}} (agent1);
        \end{tikzpicture}
        \caption{An adversary manipulates with $\delta$ the observations of the RL policy $\pi$ and its interaction with the environment.}
        \label{fig:advl_rl}
    \end{subfigure}
    \hfill 
    \caption{RL (a) vs. adversarial RL (b).}
    \label{fig:rl_adv}
\end{figure}
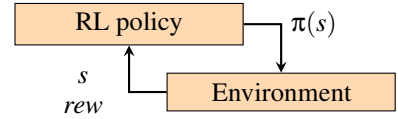
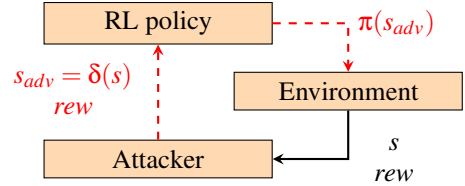

\begin{definition}[Adversarial Attack]\label{def:adv_attack}
    An \emph{adversarial attack} $\delta \colon S \rightarrow S$ maps a state $s$ to an \emph{adversarial state} $s_{adv}$ (see \cref{fig:rl_adv}). A successful adversarial attack at a given state $s$ leads to a \emph{misjudgment} of the RL policy ($\pi(s)~\neq~\pi(\delta(s))$) and an attack is $\epsilon$-bounded if $\norm{\delta(s)-s}_{\infty}\leq \epsilon$ with $l_\infty$-norm defined as $\norm{\delta(s)-s}_\infty = max_{\delta_i \in \delta} |\delta_i-s_i|$.
\end{definition}
Recall that states are $n$-dimensional vectors of features from $\mathbb{Z}^n$.
Executing a policy $\pi$ on an MDP $M$ and attacking the policy $\pi$ at each reachable state $s$ by $\delta$ yields an adversarial-induced~DTMC~$D_{adv}$.
There exist a variety of adversarial attack methods to create adversarial attacks $\delta$~\citep{DBLP:journals/tai/IlahiUQJAHN22,DBLP:conf/iclr/GleaveDWKLR20,DBLP:conf/aaai/LeeGTHS20,DBLP:conf/iccps/LeeETS21,DBLP:conf/icml/RakhshaRD0S20,DBLP:conf/sp/Carlini017}.

Our work builds upon the FGSM attack and the work of \citet{DBLP:conf/ccs/ChanWY20}.
Given the weights $\theta$ of the neural network policy $\pi$ and a loss $J(\theta,s,a)$ with state $s$ and $a \coloneqq \pi(s)$,
the FGSM, denoted as $\delta_{\FGSM} \colon S \rightarrow S$, adds noise whose direction is the same as the gradient of the loss $J(\theta,s,a)$ w.r.t the state $s$ to the state $s$ \citep{DBLP:conf/iclr/HuangPGDA17} and the noise is scaled by $\epsilon \in \mathbb{Z}$~(see \cref{eq:fgsm}).
Note that we are dealing with integer $\epsilon$-values because our states are comprised of integer features.
We specify the $\bigtriangledown$-operator as a vector differential operator.
Depending on the gradient, we either add or subtract $\epsilon$.
\begin{equation}\label{eq:fgsm}
    \delta_{\FGSM}(s) = s + \epsilon \cdot sign(\bigtriangledown_{s}J(\theta,s,a))
\end{equation}
A FGSM for feature $f_i$, denoted as $\delta_{\FGSM}^{(f_i)}(s)$, modifies only the feature $f_i$ in state $s$.
\begin{equation}\label{eq:fgsmfi}
    \delta_{\FGSM}^{(f_i)}(s) = s + \epsilon \cdot sign(\bigtriangledown_{s_{f_i}}J(\theta,s,a))
\end{equation}
We denote the set of all possible $\epsilon$-bounded attacks at state $s$ via feature $f_i$, including $\delta^{(f_i)}(s) = s$ for no attack, as $\Delta^{(f_i)}_\epsilon(s)$.

\citet{DBLP:conf/ccs/ChanWY20}\label{back:chan} first generate for all features a static reward impact (SRI) map by attacking each feature (in the case of the smallest sliding window) with the FGSM attack to measure its impact (the drop of the expected reward) offline.
A feature $f_i$ with a more significant impact indicates that changing this feature $f_i$ via $\delta_{\FGSM}^{(f_i)}$ will influence the expected discounted reward more than via another feature $f_k$ with a less significant impact.
For each feature $f_i$, this is done multiple times $N$, where each iteration executes the RL policy on the environment and attacks at every state the feature $f_i$ via the FGSM attack $\delta_{\FGSM}^{(f_i)}$.
After calculating the SRI, they use all the SRI values of the features $f_i$ to select the most vulnerable feature to attack the deployed RL policy.

\emph{Adversarial training} retrains the already trained RL policy by using adversarial attacks during training (see \cref{fig:advl_rl}) to increase the RL policy robustness \citep{DBLP:conf/icml/PintoDSG17,DBLP:journals/corr/abs-2205-14691,DBLP:conf/uai/Korkmaz21}.

\section{\uppercase{Methodology}}\label{sec:methodology}
We introduce the general adversarial setting, the property impact (PI), the property impact attack (PIA), and bounded robustness.

\subsection{Attack Setting}\label{sec:as}
We first describe our method's adversarial attack setting (adversary's goals, knowledge, and capabilities).

\textbf{Goal.} 
The adversary aims to modify the property value of the target RL policy $\pi$ in its environment (modeled as an MDP).
For instance, the adversary may try to increase the probability that the agent collides with another object (i.e. $\max_\delta P(F\text{ }\mathit{collision})$ in the adversarial-induced~DTMC).

\textbf{Knowledge.} 
The adversary that knows the weights $\theta$ of the trained policy (for the FGSM attack) and knows the MDP of the environment.
Note that we can replace the FGSM attack with any other attack. Therefore, knowing the weights of the trained policy should not be a strict constraint.

\textbf{Capabilities.} 
The adversary can attack the trained policy $\pi$ at every visited state $s$
during the incremental building process for the model checking of the adversarial-induced DTMC and after the RL policy got deployed.

\subsection{Property Impact Attack (PIA)}\label{sec:pia}
Combining adversarial RL with model checking allows us to craft adversarial property impact attacks (PIAs) that target temporal logic properties.
Our work builds upon the research of \citet{DBLP:conf/ccs/ChanWY20}.
Instead of calculating SRIs (see \cref{back:chan}), we calculate property impacts (PIs).
The PI values are used to select the feature $f_i$ with the most significant $PI$-value to attack the deployed RL policy in its environment ($f_i = \argmax_{f_i \in \mathcal{F}} PI(\pi,P(\phi),f_i,\epsilon)$). 
\begin{definition}[Property Impact]\label{def:pi}
  The \emph{property impact}  $PI \colon \Lambda \times \Theta \times \mathcal{F} \times \mathbb{Q} \rightarrow \mathbb{Q}$ quantifies the impact of an adversarial attack $\delta_{\FGSM}^{(f_i)} \in \Delta^{(f_i)}_\epsilon(s)$ via a feature $f_i \in \mathcal{F}$ on a given RL policy property $P(\phi) \in \Theta$ with $\Theta$ as the set of all possible PCTL properties for the MDP $M$.
\end{definition}
A feature $f_i$ with a more significant PI-value indicates that changing this feature $f_i$ via $\delta_{\FGSM}^{(f_i)}$ will influence the property (expressed by the property query $P(\phi)$) more than via another feature $f_k$ with a less significant PI-value.

\begin{algorithm}[t]
  \caption{Calculate the property impact (PI) for a given MDP $M$, policy $\pi$, property query $P(\phi)$, feature $f_i$, and FGSM attack strength $\epsilon$.}\label{euclid}
  \begin{algorithmic}[1]
    \Procedure{PI}{$\pi,P(\phi),f_i,\epsilon$}
    \State $r\gets property\_result(M,\pi,P(\phi))$
    \State $r_{adv} \gets adv\_property\_result(M,\pi,P(\phi),f_i,\epsilon)$
    \State \textbf{return} $|r-r_{adv}|$
    \EndProcedure
  \end{algorithmic}
\label{alg:pi}
\end{algorithm}

In \cref{alg:pi}, we explain how to calculate the PI-value for a given MDP $M$, policy $\pi$, PCTL property query $P(\phi)$, feature $f_i$, and FGSM attack $\delta_{\FGSM}^{(f_i)}$.
First, we incrementally build the induced DTMC of the policy $\pi$ and the MDP $M$ to check the property value $r$ of the policy $\pi$ via the function \emph{property\_result}.
The function \emph{property\_result} uses COOL-MC and inputs the MDP $M$, policy $\pi$, and PCTL property query $P(\phi)$ into it to calculate the probability $r$.
Second, we incrementally build the adversarial-induced DTMC $D_{adv}$ of the policy $\pi$ and the MDP $M$ with the $\epsilon$-bounded FGSM attack $\delta_{\FGSM}^{(f_i)}$ to check its probability $r_{adv}$ via the function $adv\_property\_result$. To support the building and model checking of adversarial-induced DTMCs via $adv\_property\_result$, we extend the incremental building process of COOL-MC in the following way.
For every reachable state $s$ by the policy $\pi$, the
policy $\pi$ is queried for an action $a = \pi(s)$.
In the underlying MDP, only states $s$ that may be reached via that action $a$ are expanded.
The resulting model is fully probabilistic, as no action choices are left open. It is, in fact, the Markov chain induced by the original MDP $M$ and the policy $\pi$.
An adversary can now inject adversarial attacks $\delta(s)$ at every state $s$ that gets passed to the policy $\pi$ during the incrementally building process~\citep{DBLP:conf/nips/0001CX0LBH20}.
This may lead to the effect that the policy $\pi$ makes a misjudgment $(\pi(s)\neq\pi(\delta(s))$ and results into an adversarial-induced DTMC $D_{adv}$.
This allows us to model check the adversarial-induced DTMCs $D_{adv}$ to gain the adversarial probability $r_{adv}$.
Finally, we measure the property impact value by measuring the absolute difference between $r$ and $r_{adv}$.

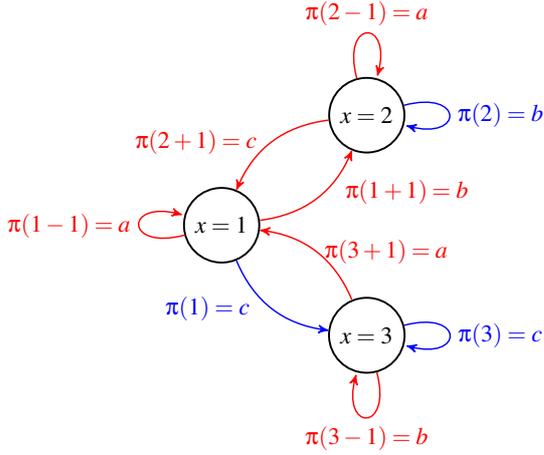
\begin{figure}[tbp]
\scalebox{0.9}{
\begin{tikzpicture}[->, >=stealth', auto, semithick, node distance=3cm]
    \tikzstyle{every state}=[fill=white,draw=black,thick,text=black,scale=1]
    \node[state]    (A)                    {$x=1$};
    \node[state]    (B)[above right of=A, yshift=-.5cm]   {$x=2$};
    \node[state]    (C)[below right of=A, yshift=.5cm]   {$x=3$};
    \path
    (A) edge[loop left, red]     node{$\pi(1-1)=a$}         (A)
        edge[red, bend right]     node[below right,yshift=0.4cm,xshift=0.3cm]{$\pi(1+1)=b$}     (B)
        edge[blue, bend right]    node[left,xshift=-0.2cm]{$\pi(1)=c$}      (C);
    \path
    (B) edge[loop above, red]     node[]{$\pi(2-1)=a$}         (B)
        edge[loop right, blue]     node{$\pi(2)=b$}         (B)
        edge[red, bend right]  node[left,xshift=-0.1cm]{$\pi(2+1)=c$}     (A);
    \path
    (C) edge[loop below, red]     node[below]{$\pi(3-1)=b$}         (C)
        edge[loop right, blue]     node{$\pi(3)=c$}         (C)
        edge[red,  bend right]  node[right]{$\pi(3+1)=a$}     (A);
    \end{tikzpicture}
}
\caption{($\epsilon,\alpha$)-robustness checking for an MDP with the state set $S=\{x \in [0,4]\}$, action set $Act=\{a,b.c\}$, trained policy $\pi$, adversarial attack strength $\epsilon=1$ ($\Delta_{\epsilon=1}(s) = \{s+1,s-1,s\}$), and property $P(F\text{ }x=2)=0$ (for original blue policy $\pi$). The blue induced DTMC is the original one, and the MDP (blue and red) is the adversarial MDP representing the permissive policy $\Omega$.
For the permissive policy $\Omega$, we get $P^{max}(F\text{ }x=2)=1$ which results in a $P^{max}(F\text{ }x=2)-P(F x = 2)=1$ and indicate that $\pi$ is not robust for, for example, $\alpha=0$ and $\epsilon=1$.
We can extract the optimal attack set $\{\delta(s)=s+1 \text{ at state }x=1\}$.
}
\label{fig:permissive_building}
\end{figure}

\subsection{RL Policy Robustness}
A trained RL policy $\pi$ can be robust against an $\epsilon$-bounded PIA that attacks a temporal logic property $P(\phi)$ via feature $f_i$ ($PI(\pi,P(\phi),f_i, \epsilon) = 0$).
However, this is a weak statement about robustness since there still exist multiple adversarial attacks $\delta^{(f_i)}(s)$ with $\norm{\delta^{(f_i)}(s)-s}_{\infty} \leq \epsilon$ generated by other attacks, such as the method from \citet{DBLP:conf/sp/Carlini017}.

Given a fixed policy $\pi$ and a set of attacks $\Delta^{(f_i)}_\epsilon(s)$, we generate a \emph{permissive policy} $\Omega$. Applying this policy $\pi$ in the original MDP $M$ generates a new MDP $M'$ that describes all potential behavior of the agent under the attack.
\begin{definition}[Behavior under attack]\label{def:per}
    A permissive policy~$\Omega \colon S \rightarrow 2^\textsf{Act}$ selects, at every state $s$, all actions that can be queried via $\Delta^{(f_i)}_s(s)$.
    We consider $\Omega(s) = \bigcup_{\delta^{(f_i)}_i \in \Delta^{(f_i)}_s(s)} \pi(\delta^{(f_i)}_i(s))$ with $\pi(\delta^{(f_i)}_i(s)) \in Act(s)$.    
\end{definition}
Applying a permissive policy to an MDP does not necessarily resolve all nondeterminism, since more than one action may be selected in some state(s).
The induced model is then (again) an MDP. 
We are able to apply model checking, which typically results in best- and worst-case probability bounds $P^{max}(\phi)$ and $P^{min}(\phi)$ for a given property query~$P(\phi)$.

We use the induced MDP to model check the \emph{robustness} (see \cref{def:robustness}) against every possible $\epsilon$-bounded attack $\delta^{(f_i)}(s)$ for a trained RL policy $\pi$ in its environment and bound the robustness to an $\alpha$-threshold (property impacts below a given threshold $\alpha$ may be acceptable).
\begin{definition}[Bounded robustness]\label{def:robustness}
A policy $\pi$ is called \textit{robustly bounded} by $\epsilon$ and $\alpha$ ($\epsilon,\alpha$-robust) for property query $\phi$ if it holds that
\begin{equation}\label{eq:robustness}
    |P^{*}(\phi) - P(\phi)| \leq \alpha
\end{equation}
for all possible $\epsilon$-bounded adversarial attacks $\Delta^{(f_i)}_\epsilon(s)$ at every reachable state $s$ by the permissive policy $\Omega$.
We define $\alpha \in \mathbb{Q}$ as a threshold (in this paper, we focus on probabilities and therefore $\alpha \in [0,1]$).
$ |P^{*}(\phi) - P(\phi)|$ stands for the largest impact of a possible attack.
We denote $P^{*}$ as $P^{max}$ or $P^{min}$ depending if the attack should increase ($P^{max}$) or decrease ($P^{min}$) the~probability.
\end{definition}

By model checking the robustness of the trained RL policies (see \cref{fig:permissive_building}), it is possible to extract for each state $s$ the adversarial attack $\delta^{(f_i)}$ that is part of the most impactful attack and use the corresponding attack as soon as the state gets observed by the adversary.
This is possible because the underlying model of the induced MDP allows the extraction of the state and action pairs ($s, a_{adv}$) that lead to the wanted property value modification ($a_{adv}\coloneqq \pi(\delta^{(f_i)}(s))$).


\section{\uppercase{Experiments}}
We now evaluate our PI method, property impact attack (PIA), and robustness checker method in multiple environments.
The experiments are performed by initially training the RL policies using the deep Q-learning algorithm~\citep{mnih2013playing}, then using the trained policies to answer our research questions.
First, we compare our PI-method with the SRI-method~\citep{DBLP:conf/ccs/ChanWY20}.
Second, we use PIAs to attack policy properties.
Third, we discuss the limitations of PIAs.
Last but not least, we show that it is possible to make trained policies more robust against PIAs by using adversarial training.

\subsection{Setup}
We now explain the setup of our experiments.

\begin{figure}[tbp]
  \centering
   {\epsfig{file = 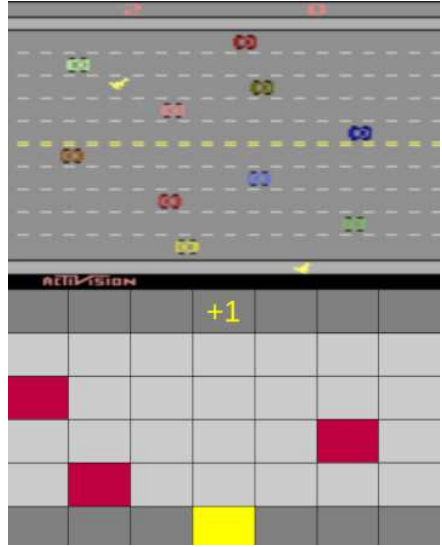, width = 6cm}}
  \caption{A comparison between the Atari 2600 Freeway game (top) and our abstracted version (bottom).}
  \label{fig:freeway}
 \end{figure}

\textbf{Environments.}\label{pg:envs} We used our proposed methods in a variety of environments (see \cref{fig:freeway}, \cref{fig:different_properties}, and \cref{tab:verification}). We use the \emph{Freeway} (for a fair comparison between the SRI  and PI method) and the \emph{Taxi}.
Additionally, we use the environments \emph{Collision Avoidance}, \emph{Stock Market}, and \emph{Smart Grid} (see \cref{sec:environments} for more details).

\emph{Freeway} is an action video game for the Atari 2600. A player controls a chicken (up, down, no operation) who must run across a highway filled with traffic to get to the other side. Every time the chicken gets across the highway, it earns a reward of one. An episode ends if the chicken gets hit by a car or reaches the other side.
Each state is an image of the game's state.
Note that we use an abstraction of the original game (see \cref{fig:freeway}), which sets the chicken into the middle column of the screen and contains fewer pixels than the original game, but uses the same reward function and actions.

The \emph{taxi} agent has to pick up passengers and transport them to their destination without running out of fuel. The environment terminates as soon as the taxi agent does the predefined number of jobs or runs out of fuel. After the job is done, a new guest spawns randomly at one of the predefined locations.
We define the maximal taxi fuel level as ten and the maximal number of jobs as two.
To refuel the taxi, the agent needs to drive to the gas station cell ($x=1,y=2$).
The problem is formalized as follows:
\begin{align*} 
S &= \{(x,y,Xloc,Yloc,Xdest,Ydest,\\& \qquad \qquad fuel,done,pass,jobs,done),...\}\\
Act &= \{north,east,south,west,pick\_up,drop\}\\
rew &= \begin{cases}
        0 \text{, if passenger successfully dropped.}
        \\
        21 \text{, if passenger got picked up.}
        \\
        21 + |x-Xdest| +\\ \quad |y-Ydest| \text{, if passenger on board.}
        \\
        21 + |x-Xloc| + |y-Yloc| \text {, otherwise.}
        \end{cases}
\end{align*}

\textbf{Properties.} \cref{tab:pctl_labels} presents the property queries of the policy trained by an RL agent achieves in these properties without the attack ($=$).
For example, $pass\_empty$ describes the probability of the taxi agent running out of fuel while having a passenger on board, which is $0$ for the policy without an attack.

\textbf{Trained RL policies.}
We trained deep Q-learning policies for the environments. See \cref{app:training} for a more detailed description of the RL~training.

\begin{table*}[tbp]
\centering
\scaleTable{
\begin{tabular}{@{}lllr@{}}
\toprule
\textbf{Env.}     & \textbf{Label}                         & \textbf{PCTL Property Query ($P(\phi)$)}                                                                             &  \textbf{$=$}                                                                                               \\ \midrule 
Fr                 & crossed                                & $P(F\text{ }crossed)$                                                                                    &  $1.0$      \\ \midrule
Taxi              & deadlock1                              & $P(fuel\geq4\text{ }U\text{ }(G (jobs=1 \land \lnot empty\text{ }\land pass)))$                           &  $0.0$      \\ \cmidrule{2-4}
                  & deadlock2                              & $P(fuel \geq 4\text{ }U\text{ }(G (jobs=1 \land \lnot empty \land \lnot pass)))$                          &  $0.0$      \\ \cmidrule{2-4}
                  & station\_empty                         & $P((((jobs{=}0$ $U$ $x{=}1 \land y{=}2)$ $U$ $(jobs{=}0 \land \lnot(x{=}1 \land y{=}2)))$ $U$ $empty \land jobs{=}0))$ &  $0.0$      \\ \cmidrule{2-4}
                  & $\overline{\textrm{station}}$\_empty   & $P(F \text{ } (empty \land jobs=0) \land G \lnot(x\neq1 \land y\neq 2))$                                 &  $0.0$      \\ \cmidrule{2-4}
                  & pass\_empty                            & $P(F \text{ } (empty \land pass))$                                                                       &  $0.0$      \\ \cmidrule{2-4}
                  & $\overline{\textrm{pass}}$\_empty      & $P(F \text{ } (empty \land \lnot pass))$                                                                 &  $0.0$      \\ \midrule
Coll.             & collision                              & $P(F_{\leq 100}\text{ }collision)$                                                                       &  $0.1$      \\ \midrule
SG                & blackout                               & $P(F_{\leq100}\text{ }blackout)$                                                                         &  $0.2$      \\ \midrule
SM                & bankruptcy                             & $P(F\text{ }bankruptcy)$                                                                                 &  $0.0$      \\ \bottomrule
\end{tabular}
}
\caption{PCTL property queries, with their labels and the original result of the property query without an attack ($=$).
$Fr$ stands for \emph{Freeway}, $Coll.$ stands for \emph{Collision Avoidance}, $SG$ for \emph{Smart Grid}, and $SM$ for \emph{Stock Market}.}
\label{tab:pctl_labels}
\end{table*}

\textbf{Technical setup.} All experiments were executed on an NVIDIA GeForce GTX 1060 Mobile GPU, 16 GB RAM, and an Intel(R) Core(TM) i7-8750H CPU~@~2.20GHz~x~12.
For model checking, we use Storm 1.7.1 (dev).

\subsection{Analysis}
We now answer our research questions.

\emph{Does the PI method have the same behavior as the related SRI method?}
We showcase that our PI approach yields similar results to the empirical SRI approach~\citep{DBLP:conf/ccs/ChanWY20} in the Freeway environment.
We chose the Freeway environment since this environment was also used by ~\citet{DBLP:conf/ccs/ChanWY20}.

\begin{figure}[tbp]
  \centering
   {\epsfig{file = 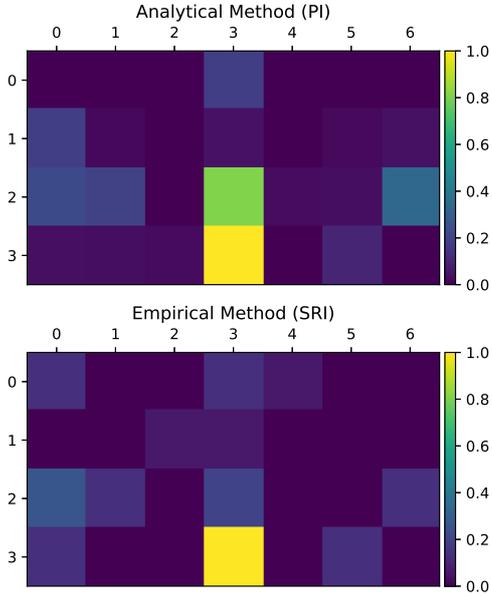, width = 7cm}}
  \caption{This plot contains the feature impacts (normalized between $0$ and $1$) for the PI, and SRI concerning the total expected reward (sample size $N = 300$) approaches with~$\epsilon=1$ of the policy $\pi_A$ in the Freeway environment (street pixels without the sidewalks).}
  \label{fig:maps}
\end{figure}

To compare both approaches, we can use the reward function (a reward of one if the player crosses the street) to express the expected reachability probability of crossing the street (see label \emph{crossed} in \cref{tab:pctl_labels}).
We then sampled for each feature the SRI value multiple times ($N=300$) to generate the SRI map  (see \cref{fig:maps}).
The PI-values are calculated by \cref{alg:pi} with the property query \emph{crossed} and are used to generate the PI map (see \cref{fig:maps}).
In both cases, we used an $\epsilon=1$.

\cref{fig:maps} shows both approaches' feature impact maps (for each game pixel, a value).
In both maps, the most impactful features (pixels) concerning the total expected reward lie on the chicken's path, which is also the result of ~\citet{DBLP:conf/ccs/ChanWY20}.

\begin{figure}[tbp]
  \centering
   \scalebox{0.64}{{\epsfig{file = 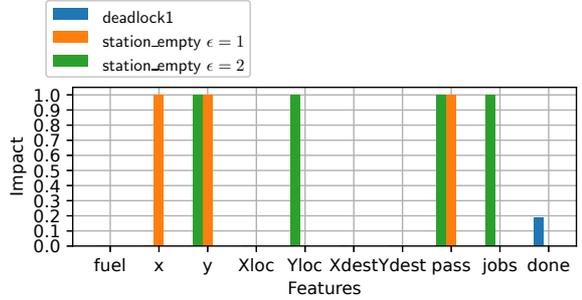}}}
  \caption{Taxi environment. This diagram plots different advanced property impacts of different PIAs. The original property values (without an attack) are all zero.}
  \label{fig:different_properties}
 \end{figure}

\emph{Can the PI method generate different property impacts for different advanced property queries?}
We now show that PI is suited to measure the property impact for properties that can not be expressed by rewards which we call here \emph{advanced property queries} (see \cref{fig:different_properties}).

To make the interpretation of advanced properties more straightforward, we focus on the Taxi environment instead of the Freeway environment and use the advanced property queries \emph{deadlock1} and \emph{station\_empty}.
Advanced property queries contain, for example, the U-operator (\cref{def:pctl}), which allows the adversary to make sure that certain events happen before other~events.

\cref{fig:different_properties} shows the property impact of each attack on the policy and different $\epsilon$-bounded attacks.
By attacking the \emph{done} feature via an PIA (with $\epsilon=1$), it is possible to drive the taxi around without running out of fuel and not finishing jobs while having a passenger on board (deadlock1).
\cref{fig:different_properties} also shows that it is possible to let the taxi drive first to the gas station and let it run out of fuel afterwards~(station\_empty).
We observe that for different $\epsilon$-bounds, PIAs have different impacts via features on the temporal logic properties (see station\_empty in \cref{fig:different_properties}).

\begin{table*}[tbp]\centering
\scaleTable{\input{table2}
}
\caption{Impact* stands for the optimal adversarial attack impact ($|P^{max}-P|$) via the feature specified in \emph{Features}, $P^{max}$ for the maximal probability $P^{max}(\phi)$ with an attack, $P$ for the original probability $P(\phi)$ (without an attack), Time in seconds, C for \emph{Collision Avoidance}, SG for \emph{Smart Grid}, SM for \emph{Stock Market}, Baseline is a standard FGSM attack on the whole observation. We observe that PIAs perform similarly to FGSM attacks for advanced properties. Our robustness checker shows that our generated PIAs are not necessarily optimal but that they can still modify the targeted property. All our attacks are $\epsilon$-bounded for a fair comparison.}
\label{tab:verification}
\end{table*}

\emph{What are the limitations of PIAs?}
We now analyze the limitations of PIAs and compare them with the FGSM attack (baseline) and the robustness checker.
For each experiment, we $\epsilon$-bounded all the generated attacks for a fair comparison.
We mainly focus on selected properties from the taxi environment but also include other environments (see~\cref{tab:verification}).

\cref{tab:verification} shows that PIAs, in comparison to FGSM attacks, have similar impacts on temporal logic properties (compare \emph{impact} columns of PIA and FGSM).
For temporal logic properties where some correct decision-making is still needed, PIAs perform better than the FGSM attack (for instance, \emph{pass\_empty}).
However, PIAs do not necessarily create a maximal impact on the property values like the robustness checker method (compare \emph{PIA impact} with~\emph{Impact*}).

After observing the results of the three methods (PIA, FGSM, robustness checker), we can summarize.
By verifying the robustness of the trained RL policies, the adversary can already extract for each state the optimal adversarial attack that is part of the most impactful attack.
Since PIAs build induced DTMCs and the robustness checker induced MDPs, PIAs are suited for MDPs with more states and transitions before running out of memory \citep[see][for more details about the limitations of model checking RL policies]{DBLP:conf/setta/Gross22} .

\emph{Does adversarial training make trained RL policies more robust against PIAs?}
\cref{fig:different_properties} shows that an adversarial attack (bounded by $\epsilon=1$) on feature $done$ can bring the taxi agent into a deadlock and lets it drive around after the first job is done ($deadlock1 = 0.19$).
To protect the RL agent from this attack, we trained the RL taxi policy over 5000 additional episodes via adversarial training by using our method PIA on the done feature to make the policy more robust against this deadlock attack.
The adversarial training improves the feature robustness for the done feature ($0$) but deteriorates the robustness for the other features (all other feature PI-values: $1$).
That agrees with the observation that adversarially trained RL policies may be less robust to other types of adversarial attacks~\citep{DBLP:conf/nips/0001CX0LBH20,korkmaz2021adversarial,DBLP:conf/aaai/Korkmaz22}.
To summarize, adversarial training can improve the RL policy robustness against specific PIAs but may also deteriorate the robustness against other~PIAs.

\section{\uppercase{Conclusion}}
We presented an analytical method to measure the adversarial attack impact on RL policy properties.
This knowledge can be used to craft fine-grained property impact attacks (PIAs) to modify specific values of temporal RL policy properties.
Our model checking method allows us to verify if a trained policy is robust against $\epsilon$-bounded PIAs.
A learner can use adversarial training in combination with the PIA to obtain more robust policies against specific PIAs.

For future work, it would make sense to combine the current research with countermeasures~\citep{DBLP:journals/corr/abs-1710-00814,DBLP:conf/dsc/XiangNLCH18,DBLP:conf/nips/HavensJS18}.
Furthermore, it would be interesting to analyze adversarial multi-agent reinforcement learning~\citep{DBLP:conf/amcc/FiguraK021,zeng2022resilience} in combination with model~checking.
Interpretable Reinforcement Learning~\citep{DBLP:conf/smc/DavoodiK21} can further use the impact results to interpret trained RL policies.

\bibliographystyle{apalike}
{\small
\bibliography{main}}
\appendix
\begin{table*}[tbp]\centering
\scaleTable{
\begin{tabular}{@{}llrrrrrr@{}}
\toprule
\textbf{Policy}  & \textbf{Env.}         & \textbf{Layers} & \textbf{Neurons} & \textbf{LR}     & \textbf{Batch} & \textbf{Episodes} & \textbf{Reward}\\ \midrule
$\pi_A$ & Freeway                & 4                & 512               & 0.0001 & 100        & 10000   & 0.96     \\ \midrule
$\pi_B$ & Taxi                & 4                & 512               & 0.0001 & 100        & 60000   & -1220.49      \\ \midrule
$\pi_C$ & Collision Avoidance                & 4                & 512               & 0.0001 & 100        & 5000   & 9817.00      \\ \midrule
$\pi_D$ & Smart Grid                & 4                & 512               & 0.0001 & 100        & 5000   & -1000.10      \\ \midrule
$\pi_E$ & Stock Market              & 4                & 512               & 0.0001 & 100        & 10000   & 326.00      \\ \bottomrule
\end{tabular}
}
\caption{$Reward$ specifies the best reward sliding window over $100$ episodes while training, $LR$ for learning rate, $Batch$ for batch size.}
\label{tab:trained_policies}
\end{table*}
\section{Additional Environments}\label{sec:environments}

\textbf{Collision avoidance.} This is an environment that contains one agent and two moving obstacles in a two-dimensional grid world. The environment terminates as soon as a collision between the agent and obstacle 1 happens. The environment contains a slickness parameter, which defines the probability that the agent stays in the same cell.
\begin{gather*}
S = \{(x,y,obs1\_x,obs1\_y,obs2\_x,obs2\_y,done),...\} \\
Act = \{north,east,south,west\}\\
rew = \begin{cases}
        0 \text{, if collision with obstacle 1}
        \\
        100 \text {, otherwise}
        \end{cases}
\end{gather*}

\textbf{Smart grid.} In this environment, a controller controls renewable and non-renewable energy production distribution.
The objective is to minimize non-renewable energy production by using renewable technologies.
If the energy consumption exceeds production, it leads to a blackout.
Furthermore, if there is too much energy in the electricity network, the energy production shuts down.
\begin{gather*}
S = \{(energy,blackout,renewable,\\non\_renewable,consumption),...\} \\
Act = \{increase\_non\_renewable,\\increase\_non\_renewable,\\decrease\_renewable, decrease\_both\}\\
rew = \begin{cases}
        - max(no\_renewable-renewable,0),\\
        \text{if no blackout.}
        \\
        - 1000 \text {, otherwise}
        \end{cases}
\end{gather*}
\textbf{Stock market.} This environment is a simplified version of a stock market simulation. The agent starts with an initial capital and has to increase it through buying and selling stocks without running into bankruptcy.

\begin{gather*}
S = \{(buy\_price, sell\_price, capital,stocks,\\last\_action\_price),...\} \\
Act = \{buy, hold, sell\}\\
rew = \begin{cases}
        \text{max(capital - initial capital, 0), if hold.}
        \\
        max(floor(\frac{capital}{buy\_price}),0)\text {, if buy.}
        \\
        max(capital+\text{number of stocks}\\\text{times sell\_price - initial capital, 0), if sell.}
        \\
        \end{cases}
\end{gather*}

\section{Deep RL Training}
\label{app:training}
The training parameters of our trained policies can be found in \cref{tab:trained_policies}.
For all training runs we set the seed of numpy, PyTorch and Storm to 128.
All RL policies were trained with the standard deep Q-learning algorithm \citep{mnih2013playing}. With $\epsilon=1$, $\epsilon_{decay}=0.99999$ ($\epsilon_{decay}=0.9999$ for freeway and Collision Avoidance), $\epsilon_{min}=0.1$, $\gamma=0.99$, and a target network replacement of 100.

\section{New COOL-MC Extensions}
COOL-MC \citep{DBLP:conf/setta/Gross22} provides a framework for connecting state-of-the-art (deep) reinforcement learning (RL) with modern model checking. In particular, COOL-MC extends the OpenAI Gym to support RL training on PRISM environments and allows verification of the trained RL policies via the Storm model checker \citep{DBLP:journals/sttt/HenselJKQV22}.
COOL-MC is available on GitHub (\url{https://github.com/LAVA-LAB/COOL-MC}).

We extended COOL-MC with an adversarial RL that allows the impact measurement of adversarial attacks on RL policy properties.
Furthermore, we extended COOL-MC to support our proposed robustness checker.
Both extensions can be found in the branch \emph{mc\_pia} in the GitHub Repository (\url{https://github.com/LAVA-LAB/MC_PIA}).


\end{document}

%% file: table2.tex
\begin{tabular}{@{}llllcrrrrcrrcrr@{}}
\toprule
\multicolumn{4}{c}{{\textbf{Setup}}}                                                                   && \multicolumn{4}{c}{\textbf{Robustness Checker}}                            && \multicolumn{2}{c}{\textbf{PIA}}    && \multicolumn{2}{c}{\textbf{Baseline (FGSM)}}    \\ \cmidrule{1-4}\cmidrule{6-9}\cmidrule{11-12}\cmidrule{14-15}
\textit{Env.}  & {\textit{Features}} & {\textbf{$\epsilon$}} & \textit{Property Query}                 && {\textbf{$P^{max}$}} & {\textbf{$P$}} & {\textit{Impact*}} & \textit{Time} && {$Impact$} & \textit{Time}          && {\textbf{$Impact$}} & \textit{Time}             \\ \cmidrule{1-4}\cmidrule{6-9}\cmidrule{11-12}\cmidrule{14-15}
Taxi           & {done}              & {1}                   & deadlock1                               && {0.44}             & {0.0}          & {0.44}           & 9             && {0.19}   & 20                     && {0.00}              & 6                         \\ 
               & {done}              & {1}                   & deadlock2                               && {0.00}               & {0.0}          & {0.00}             & 9             && {0.00}     & 20                     && {0.00}              & 6                         \\ 
               & {fuel}              & {2}                   & pass\_empty                             && {1.00}               & {0.0}          & {1.00}             & 25            && {0.25}     & 20                     && {0.00}              & 6                         \\ 
               & {y}                 & {2}                   & $\overline{\textrm{pass}}$\_empty       && {1.00}               & {0.0}          & {1.00}             & 27            && {1.00}     & 20                     && {1.00}              & 6                         \\ 
               & {x}                 & {1}                   & $\textrm{station}$\_empty               && {1.00}               & {0.0}          & {1.00}             & 24            && {1.00}     &      6                 && {1.00}              & 6                         \\ 
               & {x}                 & {1}                   & $\overline{\textrm{station}}$\_empty    && {1.00}               & {0.0}          & {1.00}             & 30            && {1.00}     & 6                      && {1.00}              & 6                         \\ \cmidrule{1-4}\cmidrule{6-9}\cmidrule{11-12}\cmidrule{14-15}
C              & {obs1\_x}           & {1}                   & collision                               && {0.87}               & {0.1}          & {0.86}             & 65            && {0.46}     & 213                    && {0.87}              & 211                       \\ \cmidrule{1-4}\cmidrule{6-9}\cmidrule{11-12}\cmidrule{14-15}
SG             & {non\_renewable}    & {1}                   & blackout                                && {0.97}               & {0.2}          & {0.95}             & 2             && {0.39}     & 2                      && {0.98}              & 2                         \\ \cmidrule{1-4}\cmidrule{6-9}\cmidrule{11-12}\cmidrule{14-15}
SM             & {sell\_price}       & {1}                   & bankruptcy                              && {0.81}               & {0.0}          & {0.81}             & 15            && {0.08}     & 20                     && {0.00}              & 4                         \\ \bottomrule
\end{tabular}